\documentclass[conference]{IEEEtran}
\IEEEoverridecommandlockouts
\usepackage{cite}
\usepackage{float}
\restylefloat{table}
\usepackage{amsmath,amssymb,amsfonts}
\usepackage{algorithmic}
\usepackage{graphicx}
\usepackage{textcomp}
\usepackage{hyperref}
\usepackage{xcolor}
\def\BibTeX{{\rm B\kern-.05em{\sc i\kern-.025em b}\kern-.08em
    T\kern-.1667em\lower.7ex\hbox{E}\kern-.125emX}}
\begin{document}

\title{AI based Presentation Creator With Customized Audio Content Delivery\\

}

\author{\IEEEauthorblockN{1\textsuperscript{st} Muvazima Mansoor}
\IEEEauthorblockA{\textit{ECE} \\
\textit{PES University}\\
Bengaluru, India \\
muvazima99@gmail.com}
\and
\IEEEauthorblockN{2\textsuperscript{nd}Srikanth Chandar}
\IEEEauthorblockA{\textit{ECE} \\
\textit{PES University}\\
Bengaluru, India \\
srikanth.chandar@gmail.com}
\and
\IEEEauthorblockN{3\textsuperscript{rd} Ramamoorthy Srinath}
\IEEEauthorblockA{\textit{CSE} \\
\textit{PES University}\\
Bengaluru, India \\
ramamoorthysrinath@gmail.com}
}

\maketitle

\begin{abstract}
In this paper we propose an architecture to a solve a novel problem statement that has stemmed more so in the recent times with an increase in demand for virtual content delivery due to the COVID-19 pandemic. All educational institutions, work-places, research centres, etc. are trying to bridge the gap of communication during these socially distanced times with the use of online content delivery. The trend now is to create presentations, and then subsequently deliver the same using various virtual meeting platforms. The time being spent in such creation of presentations and delivering is what we try to reduce and eliminate through this paper which aims to use Machine Learning (ML) algorithms and Natural Language Processing (NLP) modules to automate the process of creating a slides based presentation from a document, and then use State-of-the-art voice cloning models to deliver the content in the desired author's voice. We consider a structured document such as a research paper to be the content that has to be presented. The research paper is first summarized using BERT summarization techniques and condensed into bullet points that go into the slides.  Tacotron inspired architecture with Encoder, Synthesizer, and a Genarative Adversarial Network (GAN) based Vocoder, is used to convey the contents of the slides in the author’s voice (or any customized voice).
The world is facing a pandemic and the people had to make significant changes in their lifestyles to adapt to it. Almost all learning has now been shifted to online mode, and working professionals are now working from the comfort of their homes. Due to the current situation, teachers and professionals have shifted to presentations to help them in imparting information. In this paper, we aim to reduce the considerable amount of time that is taken in creating a presentation by automating this process and subsequently delivering this presentation in a customized voice, using a content delivery mechanism that can clone any voice using a short audio clip. 

\end{abstract}


\section{Introduction}
A system/product aiming to work on the exact same application is non-existent at the moment. There have been subsets to the application we propose that has garnered significant attention over the last few years. The ideas behind techniques and the algorithms used in summarization models is not entirely new. However, the use of such algorithms in the context of summarizing a research paper and subsequently generating a slides based presentation is unheard of. 
The concept of voice cloning is one that is gaining traction, but none that we know use the idea in the context of automating the process of virtual content delivery.
Multiple speech synthesis startups like Lyrebird and Sonantic have obtained sizable grants and investments. Lyrebird aims to offer their voice cloning API so that third parties can make use of the audio mimicry technology for their own needs. Whereas, sonantic aims to use their voice cloning feature in video games. 
The objective of our project is to implement the voice cloning feature to read out a presentation created by summarizing a research paper using a custom voice.

\section{Structure}
The project involves 4 sub-problems:

\begin{itemize}
\item Identification of sup-topics from the paper and converting these topics to hierarchical bullet points which can go on each slide of a presentation.
\item Content Generation from these points in each slide.
\item Voice recognition- mimic the style and tone of a chosen voice..
\item Present the above content by using a customized text - to - content delivery mechanism. 

\end{itemize}

\begin{figure}
\centerline{\includegraphics{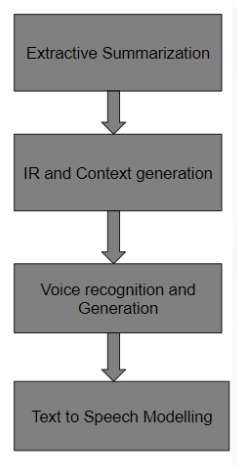}}
\caption{Structure Flowchart}
\label{fig1}
\end{figure}

\section{Past Work}
\subsection{Transfer Learning from Speaker Verification to Multispeaker Text-To-Speech Synthesis}
Ye Jia, et al. describe a text-to-speech synthesis using multiple neural networks that can generate speech audio in multiple voices, including those that are not seen during training. \cite{b1}

The system consists of three independently trained components: 

Encoder: It is used to generate a fixed-dimensional embedding vector from a few seconds of reference speech. It is trained on a speaker verification task using the LibreSpeech dataset that consists of an untranscribed noisy speech from hundreds of speakers.

Synthesizer: based on the speaker embedding derived from the encoder, the synthesizer generates a Mel spectrogram from the text. The synthesizer model is based on Tacotron 2. \cite{b2}

Vocoder: it converts the Mel spectrogram generated by the synthesizer into waveform samples in the time domain. The SV2TTS model uses a WaveNet-based vocoder that is auto-regressive. They demonstrate that the model is capable of transferring the speaker variability knowledge that is learned by the encoder to the multispeaker text to speech task and it is able to generate natural speech from speakers that are not seen during training. 

The authors mention that a large and diverse speaker dataset that is used for training the encoder is very important in order to obtain the best performance. Finally, they show that speaker embeddings that were randomly sampled can be used to generate the voice of new speakers not seen during training which indicates that the trained model has learned speaker representation of good quality.

The SV2TTS model is capable of generating realistic speech from speakers unseen in the training set, implying that the model has learned to utilize a realistic depiction of the space of speaker variation. \cite{b6}

The SV2TTS model does not attain human-level naturalness, despite the use of a WaveNet vocoder. This is due to the added difficulty of generating speech for different types of speakers given very fewer data per speaker, and the use of low-quality datasets.\cite{b7}
\subsection{Tacotron: Towards End-to-End Speech Synthesis}
A text-to-speech model usually contains stages like an audio synthesis module, an acoustic model, and a text analysis model. Constructing these models usually requires extensive domain knowledge and may contain fragile design choices. Yuxuan Wang, et al. present Tacotron, an end-to-end generative TTS model that generates speech directly from alphabets. The model can be trained from scratch using text and audio pairs with random initialization. 

Advantages of the Tacotron model:

The tacotron model reduces the need for arduous feature engineering, which may involve fragile design choices and heuristics. 

The model allows for rich conditioning on various features like speaker, language, or sentiment. This is due to the fact that conditioning can take place at the initial stages of the model rather than only on specific components. Likewise, adjustment to the new data could also be easier. 

A single model is more likely to be robust compared to a multi-stage model where every stage’s errors can multiply. 

The advantages listed above suggest that an end-to-end model could facilitate training on large amounts of noisy data that is found everywhere.
Fig.2 shows the model architecture. 
Text to speech is a large-scale inverse issue: a highly compressed text is decompressed into audio. The same text can equate to various intonations or speaking styles, which is an arduous learning task for the model. For a given input, the model must endure large disparities at the signal level. \cite{b10}
\begin{figure}[htbp]
\centerline{\includegraphics[width=95mm,scale=0.7]{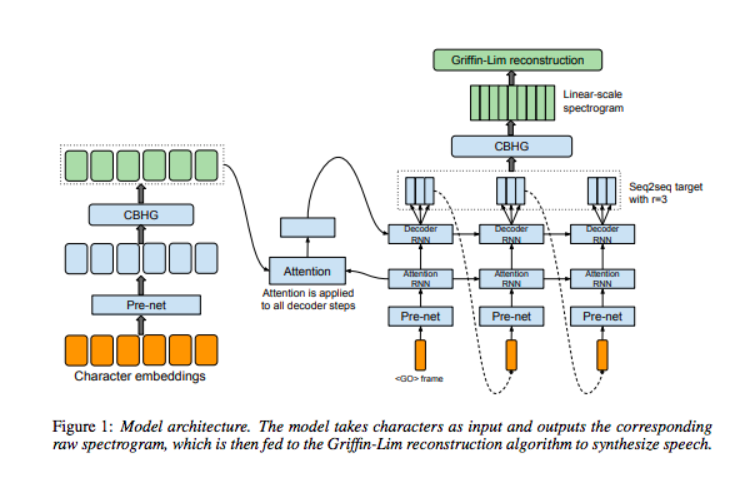}}
\caption{Tacotron Model Architecture.}
\label{fig2}
\end{figure}

\subsection{High Fidelity Speech Synthesis With Adversarial Networks}
In recent years, Generative adversarial networks have seen rapid development and have led to extraordinary developments in the generative modeling of images. Their implementation in the audio domain has received little attention. Autoregressive models like WaveNet are the most widely used in the generative modeling of audio signals. To discuss this paucity, Jeff Donahue et al. introduce Text-to-Speech using Generative Adversarial Network or GAN-TTS.\cite{b3}

Generative Adversarial Networks form a subgroup of generative models that involves the  adversarial training of two networks: 
\begin{itemize}
\item Generator: it tries to generate samples that resemble the reference distribution.
\item Discriminator: it imparts a useful gradient signal to the generator by differentiating between generated samples and the real samples.

\end{itemize}

Residual blocks are used in the model. The Convolutional layers have equal output and input channels.

GANs are capable of producing high-precision speech that sounds as natural as the ones produced by the state-of-the-art models. GANs are extremely parallelizable due to an efficient feed-forward generator. \cite{b5} Whereas, autoregressive models like WaveNet are not parallelizable. Fig 3. shows the Model Architecture. \cite{b8}

\begin{figure}[htbp]
\centerline{\includegraphics[width=95mm,scale=0.7]{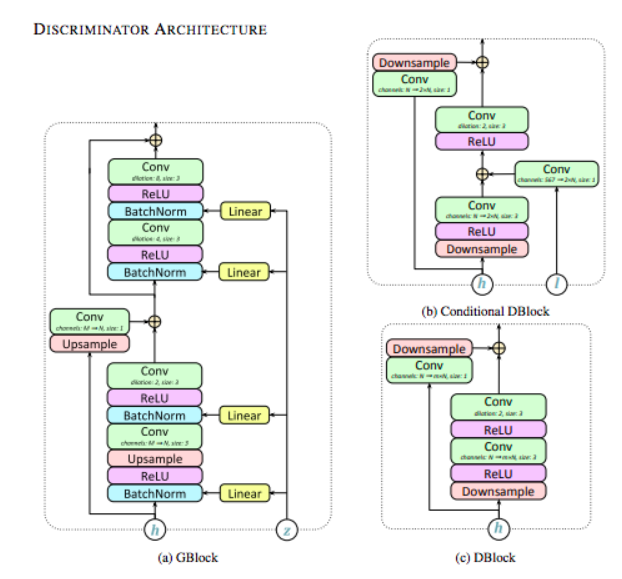}}
\caption{GAN vocoder Architecture.}
\label{fig3}
\end{figure}

\subsection{Leveraging BERT for Extractive Text Summarization on Lectures}
This paper describes a python-based RESTful service that uses Bidirectional Encoder Representations from Transformers model for text embeddings. For summary selection, The project also utilizes K Means clustering to determine the sentences that are nearest to the centroid. Apart from summarization, the project provides features for the management of lectures and summaries and supports collaboration by storing content on the cloud. \cite{b4}

There are two different types of automatic text summarization:

\begin{itemize}
\item Abstractive: Abstractive summarization resembles human summarization closely by using a vocabulary beyond the text provided. It abstracts the important points present in the text and it is usually smaller in size. Though this approach is very useful, it is arduous to produce it automatically. It requires multiple GPUs and takes many days to train.
\item Extractive: It uses only the content from the given text like the raw phrases and sentences to provide a summarization of the text.

\end{itemize}

Derek Miller uses BERT (Bidirectional Encoder Representations from Transformers) for the process of summarization. 

The unsupervised model, BERT is built on top of the Transformer architecture. It performs better than all the existing NLP models for a broad range of functions.

In other summarization models, it was not possible to obtain dynamic summary sizes. The BERT model produces sentence embeddings. These sentence embeddings can be clustered with a size of K which permits dynamic summary sizes.

BERT combines context with the most important sentences and therefore performs much better than methods like TextRank in terms of quality. \cite{b15}

\section{Methodology and System Design- Text to speech with voice cloning}
\begin{figure}[htbp]
\centerline{\includegraphics[width=95mm,scale=0.7]{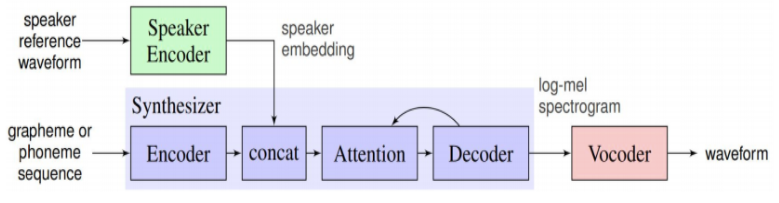}}
\caption{Architecture.}
\label{fig3}
\end{figure}

\subsection{Speaker Encoder}
The speaker encoder block is used to derive the embedding of the user’s voice from a short audio clip.
Our project uses the speaker Encoder model from SV2TTS [1].  
In SV2TTS, the speaker encoder network is trained on a speaker verification task using the LibreSpeech dataset that consists of an untranscribed noisy speech from hundreds of speakers. This allows the model to produce an embedding vector of fixed dimensions from only a few seconds of reference speech. 
Other multi-speaker speech synthesis is done by incorporating hundreds of hours of the target speaker’s voice in the training dataset. This method requires many hours of transcribed data and does not allow the fitting of new voices without retraining the computationally heavy model.

\subsection{Synthesizer} 
Based on the speaker embedding derived from the encoder, the synthesizer  		generates a Mel spectrogram from text. Our project uses the synthesizer model from Tacotron  [2]. Tacotron is an end-to-end generative TTS model that generates speech directly from alphabets. The model can be trained from scratch using text and audio pairs with random initialization. This approach does not use complex linguistic and acoustic features as input compared to other models like Deep Voice [3] and VoiceLoop [4].

\subsection{Neural Vocoder}
A vocoder converts the Mel spectrogram generated by the synthesizer into 		waveform samples in the time domain. Our project uses GAN(Generative Adversarial Networks) [5] as the vocoder. 

GANs are capable of producing high-precision speech that sounds as natural as the ones produced by the state-of-the-art models. GANs are extremely parallelizable due to an efficient feed-forward generator. Whereas, autoregressive models like WaveNet are not parallelizable.

\section{Summarization of a Research Paper and creating a presentation}

Our project uses BERT (Bidirectional Encoder Representations from Transformers) [6] summarization for the summary creation process. 
The unsupervised model, BERT is built on top of the Transformer architecture. It performs better than all the existing NLP models for a broad range of functions, including summarization.
In other summarization models, it is not possible to obtain dynamic summary sizes. The BERT model produces sentence embeddings. These sentence embeddings can be clustered with a size of K which permits dynamic summary sizes.
BERT combines context with the most important sentences and therefore performs much better than methods like TextRank in terms of quality.

Our approach is to apply BERT summarization to every section of the research paper which would go on every new page of the presentation. Each page of the presentation will also contain a hyperlink that would direct the user to the section of the research paper that has been summarized. 

To validate our results, we compared the rouge-l scores of the summary generated by the BERT summarizer by using author generated highlights as the reference summary. 

\section{Implementation}
This section provides an overview about how the various machine learning models (Summarizer and the Voice Cloning - all 3 segments- have been tied together to make a web based application). Given that this is not very research centric, this section will remain brief. The purpose of including this section is to give an idea to the reader on the implementation aspect of this research, and also how the timing constraints that one may have assumed to be a problem in the case of complex neural networks such as cloning (GAN based), is actually handled. \cite{b11} and \cite{b12}
The architecture used is that of a python implementation which combines Streamlit and Ngrok, using Google Colab as a codebase to utilize its computational power to run the ML models. Streamlit is a platform/library that enables data scripts such as the python modules used in this paper to be converted to web aplications with ease. Ngrok enables such a web application by providing a secure URL to the localhost server through any NAT or firewall. This also solves the issue in terms of security as it provides a secure authenticator key to connect to the local host (web application). The architecture as shown in Fig. 5 is connected to Google Drive as the database. The pretrained ML models- synthesizer,Encoder, and Vocoder, parsed papers, generated PPTs and audio clips are stored in Google Drive. These models are called by Google Colab as per the implementation, a web application is hosted on the local host using streamlit and ngrok is used to securely connect Colab to the local host.  \cite{b9}

\begin{figure}[htbp]
\centerline{\includegraphics[width=95mm,scale=0.7]{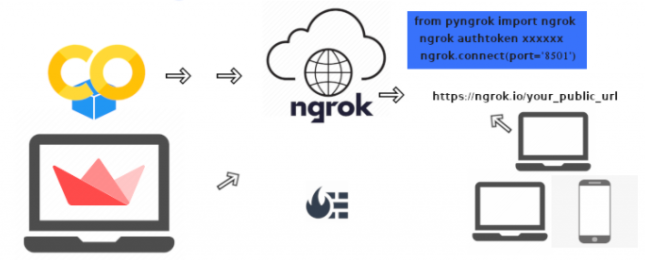}}
\caption{Web Application Architecture.}
\label{fig5}
\end{figure}

\section{Results}
The architecture proposed in this paper is one that combines different elements from methods to come up with a voice cloning approach. The encoder is from the SV2TTS model, the sythesizer has been inspired from Tacotron, and the Vocoder is the GAN based model. The architecture when combined produced voice cloning results that match the reference audio to a good degree. \cite{b13} and \cite{b14} 
To determine the accuracy of the summary generated by the BERT summarizer, we computed rouge scores of the summary generated by using the highlights provided by the author as reference summary. We also computed rouge scores of different summarization models and compared their performance with BERT summarizer. The different summarization models used are: 
\begin{itemize}
\item Abstractive summarization using TensorFlow and Keras based Neural Network. It performed poorly and gave an f-score of 0.115.
\begin{table}[H]
\centering
\begin{tabular}{l|c|c|c|}
\cline{2-4}
                                         & \textbf{f}                   & \textbf{p}          & \textbf{r}          \\ \hline
\multicolumn{1}{|l|}{\textbf{rouge -1}}  & 0.17532          & 0.14545 & 0.12188 \\ \hline
\multicolumn{1}{|l|}{\textbf{rouge -2}}  & 0.03075         & 0.06458 & 0.02072 \\ \hline
\multicolumn{1}{|l|}{\textbf{rouge -l}} & \textbf{0.11545} & 0.29515 & 0.10673 \\ \hline
\end{tabular}

\caption{Rouge scores of Abstractive summarization}
\label{table:1}
\end{table}
\item Extractive summarization using TextRank algorithm  which is an extractive and unsupervised text summarization technique. It gave an f-score of 0.1538.
\begin{table}[H]
\centering
\begin{tabular}{l|c|c|c|}
\cline{2-4}
                                        & \textbf{f}       & \textbf{p} & \textbf{r} \\ \hline
\multicolumn{1}{|l|}{\textbf{rouge -1}} & 0.19631          & 0.14545    & 0.30188    \\ \hline
\multicolumn{1}{|l|}{\textbf{rouge -2}} & 0.07453          & 0.05504    & 0.11538    \\ \hline
\multicolumn{1}{|l|}{\textbf{rouge -l}} & \textbf{0.15384} & 0.11594    & 0.22857    \\ \hline
\end{tabular}
\caption{Rouge scores of Extractive summarization using TextRank}
\label{table:2}
\end{table}
\item Extractive summarization using SVM -  we scored each sentence based on the words that overlapped between the main body of the paper and the author generated summaries. The top “n” sentences with the highest scores were picked as summary sentences. To do this, we used a Support Vector Machine Regressor to predict sentence scores based on document vectors generated by Gensim’s doc2vec function. It gave an f-score of 0.16.
\begin{table}[H]
\centering
\begin{tabular}{l|c|c|c|}
\cline{2-4}
                                        & \textbf{f}       & \textbf{p} & \textbf{r} \\ \hline
\multicolumn{1}{|l|}{\textbf{rouge -1}} & 0.20930          & 0.23076    & 0.19148    \\ \hline
\multicolumn{1}{|l|}{\textbf{rouge -2}} & 0.05882          & 0.06976    & 0.05084    \\ \hline
\multicolumn{1}{|l|}{\textbf{rouge -l}} & \textbf{0.16004} & 0.17948    & 0.14893    \\ \hline
\end{tabular}
\caption{Rouge scores of Extractive summarization using SVM}
\label{table:3}
\end{table}
\item  Extractive summarization using BERT which is an unsupervised model built on top of transformer architecture. It performed much better than the other summarization models and gave an f-score of 0.45.
\end{itemize}

\begin{table}[H]
\centering
\begin{tabular}{l|c|c|c|}
\cline{2-4}
                                        & \textbf{f}       & \textbf{p} & \textbf{r} \\ \hline
\multicolumn{1}{|l|}{\textbf{rouge -1}} & 0.442307         & 0.45098    & 0.43396    \\ \hline
\multicolumn{1}{|l|}{\textbf{rouge -2}} & 0.15686          & 0.16       & 0.15384    \\ \hline
\multicolumn{1}{|l|}{\textbf{rouge -l}} & \textbf{0.45714} & 0.45714    & 0.45714    \\ \hline
\end{tabular}
\caption{Rouge scores of Extractive summarization using BERT}
\label{table:4}
\end{table}
To compute the accuracy of the PPT generated by our application, we compared it with a PPT created by humans and treated it as a gold standard. It gave a rouge-1 score of 0.49 and rouge-l score of 0.35

\begin{table}[H]
\centering
\begin{tabular}{l|c|c|c|}
\cline{2-4}
                                        & \textbf{f}       & \textbf{p} & \textbf{r} \\ \hline
\multicolumn{1}{|l|}{\textbf{rouge -1}} & 0.49519          & 0.44557    & 0.55725    \\ \hline
\multicolumn{1}{|l|}{\textbf{rouge -2}} & 0.17657          & 0.15885    & 0.19872    \\ \hline
\multicolumn{1}{|l|}{\textbf{rouge -l}} & \textbf{0.35419} & 0.32051    & 0.39577    \\ \hline
\end{tabular}
\caption{Rouge scores of PPT generated by BERT with human generated PPT}
\label{table:5}
\end{table}

The voice cloning results can be found \href{https://docs.google.com/presentation/d/18PEQAAqLoSWyHcxaOynjk32giYDH8XFClRfUA1qnJT8/edit?usp=sharing}{here}. As is evident from the voices, the cloning module has successfully copied the fundamental properties of the voices like pitch, depth, timbre, frequency, etc. It however does not account for style specific features like accent, voice style, specific pronunciations etc. The 3 voices considered for the sake of results are the author's, and Amitabh Bachchan (a famour celebrity in Bollywood). The cloned voices can be contrasted with the original voices which may be available to notice the similarity. 
\section*{Acknowledgment}
This work was done as part of the Undergraduate Final Year Capstone Project in PES University, Computer Science Department. We acknowledge all the support from PES University, the Computer Science Department, and the Electronics and Communication Department towards this work. For this, we would like to thank Dr. Shylaja S S of PES University and the rest of the mentors from the Computer Science department for providing us this opportunity and for their continuous support. We extend our thanks to PES University, who provided us a platform that helped us to team up and pursue this project. We also thank Dr. Anuradha M, for her support.

\vspace{12pt}

\end{document}